\documentclass{ecai2012}
\usepackage{times}
\usepackage{enumerate}
\usepackage{algpseudocode}
\usepackage[ruled]{algorithm}
\usepackage{url}
\usepackage{framed}
\usepackage{amsfonts,amsmath,amsthm,amssymb}
\usepackage{graphicx}
\usepackage{url}
\usepackage{color}

\newcommand {\IE} {\ensuremath {\mathbb{E}}}

\newtheorem{dfn}{Definition}
\newtheorem{thm}{Theorem}

\newtheorem{crl}{Corollary}

\title{VOI-aware MCTS}
\author {David Tolpin \and Solomon Eyal Shimony \institute{Ben-Gurion
    University of the Negev, Israel, 
email: \{tolpin,shimony\}@cs.bgu.ac.il} }

\begin{document}

\maketitle

\begin{abstract}
UCT, a state-of-the art algorithm for Monte Carlo tree search (MCTS)
in games and Markov decision processes, is based on UCB1, a sampling
policy for the Multi-armed Bandit problem (MAB) that 
minimizes the cumulative regret.  However, search differs from MAB in
that in MCTS it is usually only the final ``arm pull'' (the actual
move selection) that collects a reward, rather than all ``arm pulls''.
In this paper, an MCTS sampling policy based on Value of Information
(VOI) estimates of rollouts is suggested. Empirical evaluation of the
policy and comparison to UCB1 and UCT is performed on random MAB instances
as well as on Computer Go.
\end{abstract}

\section{Introduction}

MCTS, and especially UCT \cite{Kocsis.uct} appears in numerous search
applications, such as \cite{Eyerich.ctp}. Although these methods are
shown to be successful empirically, most authors appear to be using
UCT ``because it has been shown to be successful in the past'', and
``because it does a good job of trading off exploration and
exploitation''. While the latter statement may be correct for the
Multi-armed Bandit problem and for the UCB1 algorithm \cite{Auer.ucb},
we argue that a simple reconsideration from basic principles can
result in schemes that outperform UCT.

The core issue is that in MCTS for adversarial search and search in
``games against nature'' the goal is typically to find the best
first action of a good (or even optimal) policy, which is closer to
minimizing the simple regret, rather than the cumulative regret
minimized by UCB1.  However, the simple and the cumulative regret
cannot be minimized simultaneously; moreover, \cite{Bubeck.pure} shows
that in many cases the smaller the cumulative regret, the greater the
simple regret.

We begin with background definitions and related work.  VOI estimates
for arm pulls in MAB are presented, and a VOI-aware sampling policy is
suggested, both for the simple regret in MAB and for MCTS.  Finally,
the performance of the proposed sampling policy is evaluated on sets
of Bernoulli arms and on Computer GO, showing the improved
performance.

\section{Background and Related Work}
\label{sec:related-work}

Monte-Carlo tree search was initially suggested as a scheme for
finding approximately optimal policies for Markov Decision Processes
(MDP).  MCTS explores an MDP by performing
\emph{rollouts}---trajectories from the current state to a state in
which a termination condition is satisfied (either the goal or a
cutoff state).

Taking a sequence of samples in order to minimize the regret of a
decision based on the samples is captured by the Multi-armed Bandit
problem (MAB) \cite{Vermorel.bandits}. In MAB, we have a set of $K$
arms. Each arm can be pulled multiple times. When the $i$th arm is
pulled, a random reward $X_i$ from an unknown stationary distribution
is encountered. In the \textit{cumulative setting}, all encountered rewards are
collected.  UCB1 \cite{Auer.ucb} was shown to be
near-optimal in this respect. UCT, an extension of UCB1 to MCTS is
described in \cite{Kocsis.uct}, and shown to outperform many state of
the art search algorithms in both MDP and adversarial search
\cite{Gelly.mogo,Eyerich.ctp}. In the \textit{simple regret setting}, the agent
gets to collect only the reward of the last pull.
\begin{dfn}
The \textbf{simple regret} of a sampling policy for MAB
is the expected difference between the best expected reward
$\mu_*$ and the expected reward $\mu_j$ of the empirically best arm
$\overline X_j=\max_i\overline X_i$:
\begin{equation}
\IE r=\sum_{j=1}^K\Delta_j\Pr(\overline X_j=\max_i\overline X_i)
\label{eqn:simple-regret}
\vspace{-8pt}
\end{equation}
where $\Delta_j=\mu_*-\mu_j$.
\end{dfn}
Strategies that minimize the simple regret are called pure exploration
strategies \cite{Bubeck.pure}. 

A different scheme for control of sampling can use the principles of
bounded rationality \cite{Horvitz.reasoningabout} and rational
metareasoning \cite{Russell.right,HayRussell.MCTS}.  In search, one
maintains a current best action $\alpha$, and finds the expected gain
from finding another action $\beta $ to be better than the current
best.

\section{Upper Bounds on Value of Information}

The intrinsic VOI $\Lambda_i$ of pulling an arm is the expected decrease
in the regret compared to selecting the best arm without pulling any arm at
all. Two cases are possible:
\begin{itemize}
\item the arm $\alpha$ with the highest sample mean $\overline
  X_\alpha$ is pulled, and $\overline X_\alpha$ becomes lower than
  $\overline X_\beta$ of the second-best arm $\beta$;
\item another arm $i$ is pulled, and $\overline X_i$ becomes higher
than $\overline X_\alpha$.
\end{itemize}
The \textit{myopic} VOI estimate is of limited applicability to
Monte-Carlo sampling, since the effect of a single sample is small,
and the myopic VOI estimate will often be zero. However, for the
common case of a fixed budget of samples per node, $\Lambda_i$ can be
estimated as the intrinsic VOI $\Lambda_i^b$ of pulling the $i$th arm
for the rest of the budget.  Let us denote the current number of
samples of the $i$th arm by $n_i$, and the remaining number of samples
by $N$:
\begin{thm} $\Lambda_i^b$ is bounded from above as
\begin{eqnarray}
\label{eqn:thm-be}
 &\Lambda_\alpha^b \le \frac {N \overline X_\beta} {N+n_\alpha}\Pr(\overline X_\alpha'\le\overline X_\beta)
    \le \frac {N \overline X_\beta} {n_\alpha} \Pr(\overline X_\alpha'\le\overline X_\beta)\\
&\Lambda_{i|i\ne\alpha}^b \le \frac{ N(1-\overline  X_\alpha)} {N+n_i}\Pr(\overline X_i'\ge\overline X_\alpha)
     \le \frac {N(1-\overline X_\alpha)} {n_i}\Pr(\overline   X_i'\ge\overline X_\alpha)\nonumber
\end{eqnarray}
where $\overline X_i'$ is the sample mean of the $i$th arm after $n_i+N$ samples.
\label{thm:be}
\end{thm}

The probabilities can be bounded from above using the
Hoeffding inequality \cite{Hoeffding.ineq}:
\begin{thm} The probabilities in equations (\ref{eqn:thm-be}) are bounded from above as
\begin{align}
  \label{eqn:probound-blnk-hoeffding}
  \Pr&(\overline X_\alpha' \le \overline X_\beta)
  \le 2\exp\left(- \varphi(n_\alpha)(\overline X_\alpha - \overline X_\beta)^2 n_\alpha
  \right)\nonumber\\
  \Pr&(\overline X_{i|i\ne\alpha}' \ge \overline X_\beta)
  \le 2\exp\left(- \varphi(n_i) (\overline X_\alpha -\overline  X_i)^2 n_i \right)
\end{align}
where $\varphi(n)=2(\frac {1+n/N} {1+\sqrt {n/N}})^2 > 1.37$.
\label{thm:hoeffding-prob-bounds}
\end{thm}
\begin{crl}
An upper bound on the VOI estimate $\Lambda_i^b$ is obtained
by substituting (\ref{eqn:probound-blnk-hoeffding}) into (\ref{eqn:thm-be}).
\begin{align}
  \label{eqn:bound-blnk-hoeffding}
  \Lambda&_\alpha^b \le \hat\Lambda_\alpha^b=\frac {2N\overline X_\beta} {n_\alpha}\exp\left(- 1.37(\overline X_\alpha - \overline X_\beta)^2 n_\alpha\right)\nonumber\\
  \Lambda&_{i|i\ne\alpha}^b\le \hat\Lambda_i^b=  \frac {2N(1-\overline  X_\alpha)} {n_i}\exp\left(- 1.37(\overline X_\alpha - \overline X_i)^2 n_i\right)
\end{align}
\label{crl:bound-blnk-hoeffding}
\end{crl}
\vspace{-24pt}

\section{VOI-based Sample Allocation}

Following the principles of rational metareasoning, for pure
exploration in Multi-armed Bandits an arm with
the highest VOI should be pulled at each step. The upper bounds
established in Corollary~\ref{crl:bound-blnk-hoeffding} can be used
as VOI estimates. In MCTS, pure exploration takes place at the first
step of a rollout, where an action with the highest utility must
be chosen. MCTS differs from pure exploration in Multi-armed Bandits
in that the distributions of the rewards are not stationary. However,
VOI estimates computed as for stationary distributions work well in
practice. As illustrated by the empirical evaluation
(Section~\ref{sec:empirical-evaluation}), estimates based on upper
bounds on the VOI result in a rational sampling policy exceeding the
performance of some state-of-the-art heuristic algorithms.

\section{Empirical Evaluation}
\label{sec:empirical-evaluation}

\subsection{Selecting The Best Arm}
\label{sec:emp-arm}
\vspace{-8pt}
\begin{figure}[h!]
\centering
\includegraphics[scale=0.53]{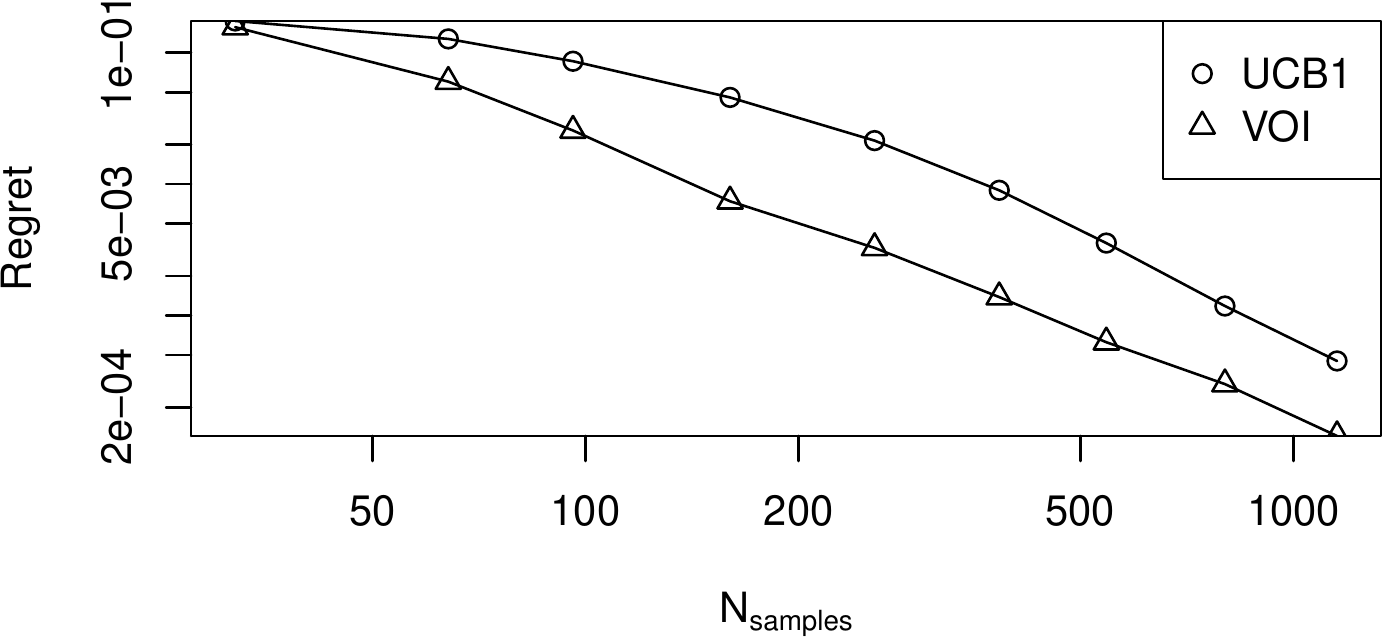}
\vspace{-16pt}
\caption{Random instances: regret vs. number of samples}
\label{fig:random-instances}
\vspace{-20pt}
\end{figure}

The sampling policies are first compared on random Multi-armed bandit 
problem instances. Figure~\ref{fig:random-instances} shows results for
randomly-generated Multi-armed bandits with 32 Bernoulli arms, with
the mean rewards of the arms distributed uniformly in the range~$[0,
  1]$, for a range of sample budgets~$32..1024$, with multiplicative
step of~$2$. The experiment for each number of samples was repeated
10000 times. UCB1 is always considerably worse than the
VOI-aware sampling policy.

\subsection{Playing Go Against UCT}
\label{sec:emp-go}

The policies were also compared on Computer Go, a  search domain
in which UCT-based MCTS has been particularly successful
\cite{Gelly.mogo}. A modified version of Pachi \cite{Braudis.pachi}, a state of the art
Go program, was used for the experiments. The UCT engine was extended
with a VOI-aware sampling policy, and a time allocation mode ensuring
that both the original UCT policy and the VOI-aware policy use the
same average number of samples per node was added. (While the UCT
engine is not the most powerful engine of Pachi, it is still a strong
player; on the other hand, additional features of more advanced
engines would obstruct the MCTS phenomena which are the subject of
the experiment.)
\begin{figure}[h!]
\centering
\includegraphics[scale=0.53]{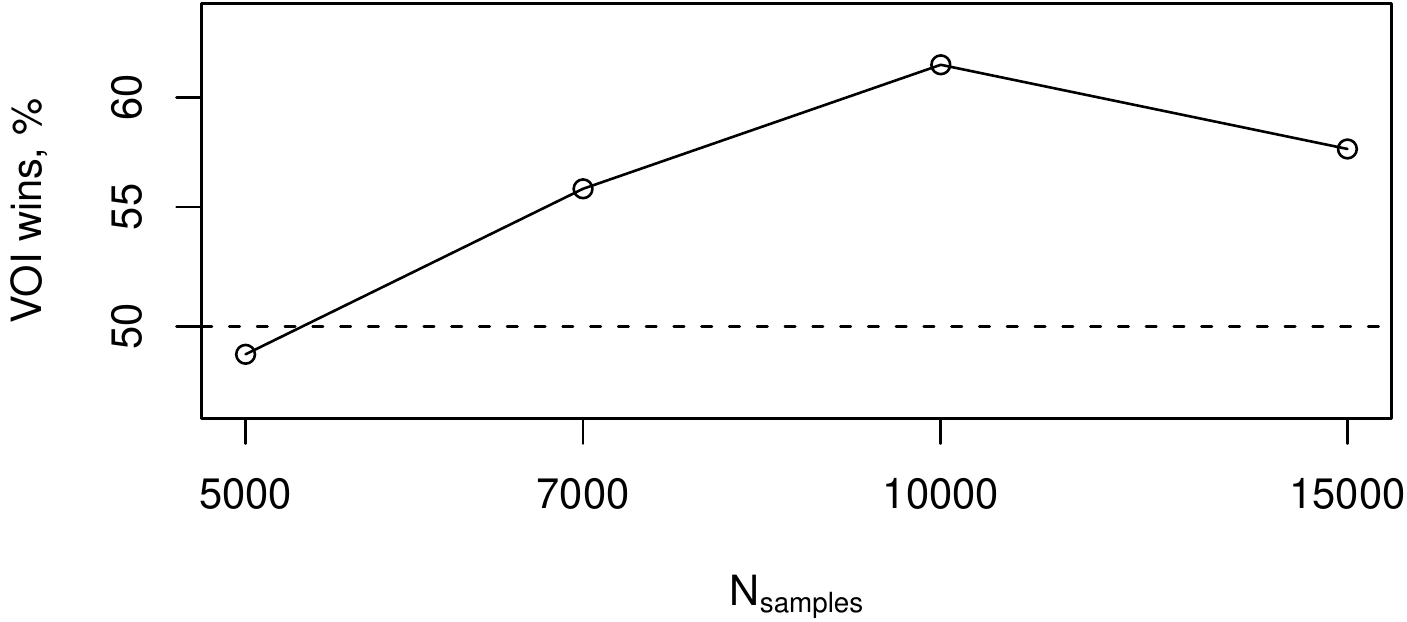}
\vspace{-16pt}
\caption{Go: winning rate --- VOI against UCT}
\label{fig:vct-against-uct}
\vspace{-20pt}
\end{figure}
The engines were compared on the 9x9 board, for 5000, 7000, 10000, and
15000 samples per ply, each experiment was repeated
1000 times. Figure~\ref{fig:vct-against-uct}
shows the winning rate of VOI against UCT vs. the number of
samples. For most numbers of samples per node, VOI outperforms UCT.

\section{Summary and Future Work}

This work suggested a Monte-Carlo sampling policy in which sample
selection is based on upper bounds on the value of
information. Empirical evaluation showed that this policy outperforms
heuristic algorithms for pure exploration in MAB, as well as for MCTS.

MCTS still remains a largely unexplored field of
application of VOI-aware algorithms. More elaborate VOI estimates,
taking into consideration re-use of samples in future search states
should be considered. The policy introduced in the paper differs from
the UCT algorithm only at the first step, where the VOI-aware
decisions are made. Consistent application of principles of rational
metareasoning at all steps of a rollout may further improve the
sampling.

\bibliographystyle{ecai2012}
\bibliography{refs}

\begin{thebibliography}{10}

\bibitem{Auer.ucb}
Peter Auer, Nicol\`{o} Cesa-Bianchi, and Paul Fischer, `Finite-time analysis of
  the {M}ultiarmed bandit problem', {\em Mach. Learn.}, {\bf 47},  235--256,
  (May 2002).

\bibitem{Braudis.pachi}
Petr Braudi\v{s} and Jean {L}oup Gailly, `Pachi: State of the art open source
  {G}o program', in {\em ACG 13}, (2011).

\bibitem{Bubeck.pure}
S{\'e}bastien Bubeck, R{\'e}mi Munos, and Gilles Stoltz, `Pure exploration in
  finitely-armed and continuous-armed bandits', {\em Theor. Comput. Sci.}, {\bf
  412}(19),  1832--1852, (2011).

\bibitem{Eyerich.ctp}
Patrick Eyerich, Thomas Keller, and Malte Helmert, `High-quality policies for
  the canadian travelers problem', in {\em In Proc. AAAI 2010}, pp. 51--58,
  (2010).

\bibitem{Gelly.mogo}
Sylvain Gelly and Yizao Wang, `Exploration exploitation in {G}o: {UCT} for
  {M}onte-{C}arlo {G}o', {\em Computer}, (2006).

\bibitem{HayRussell.MCTS}
Nicholas Hay and Stuart~J. Russell, `Metareasoning for {M}onte {C}arlo tree
  search', Technical Report UCB/EECS-2011-119, EECS Department, University of
  California, Berkeley, (Nov 2011).

\bibitem{Hoeffding.ineq}
Wassily Hoeffding, `Probability inequalities for sums of bounded random
  variables', {\em Journal of the American Statistical Association}, {\bf
  58}(301),  pp. 13--30, (1963).

\bibitem{Horvitz.reasoningabout}
Eric~J. Horvitz, `Reasoning about beliefs and actions under computational
  resource constraints', in {\em Proceedings of the 1987 Workshop on
  Uncertainty in Artificial Intelligence}, pp. 429--444, (1987).

\bibitem{Kocsis.uct}
Levente Kocsis and Csaba Szepesv{\'a}ri, `Bandit based {M}onte-{C}arlo
  planning', in {\em ECML}, pp. 282--293, (2006).

\bibitem{Russell.right}
Stuart Russell and Eric Wefald, {\em Do the right thing: studies in limited
  rationality}, MIT Press, Cambridge, MA, USA, 1991.

\bibitem{Vermorel.bandits}
Joann{\`e}s Vermorel and Mehryar Mohri, `Multi-armed bandit algorithms and
  empirical evaluation', in {\em ECML}, pp. 437--448, (2005).

\end{thebibliography}

\end{document}